\pdfoutput=1

\documentclass[
10pt, % Main document font size
a4paper, % Paper type, use 'letterpaper' for US Letter paper
oneside, % One page layout (no page indentation)
headinclude,footinclude, % Extra spacing for the header and footer
BCOR5mm, % Binding correction
]{scrartcl}

\usepackage[
nochapters, % Turn off chapters since this is an article        
beramono, % Use the Bera Mono font for monospaced text (\texttt)
eulermath,% Use the Euler font for mathematics
pdfspacing, % Makes use of pdftex’ letter spacing capabilities via the microtype package
dottedtoc % Dotted lines leading to the page numbers in the table of contents
]{classicthesis} % The layout is based on the Classic Thesis style

\usepackage{arsclassica} % Modifies the Classic Thesis package

\usepackage[T1]{fontenc} % Use 8-bit encoding that has 256 glyphs

\usepackage[utf8]{inputenc} % Required for including letters with accents

\usepackage{graphicx} % Required for including images
\graphicspath{{Figures/}} % Set the default folder for images

\usepackage{enumitem} % Required for manipulating the whitespace between and within lists

\usepackage{lipsum} % Used for inserting dummy 'Lorem ipsum' text into the template

\usepackage{subfig} % Required for creating figures with multiple parts (subfigures)

\usepackage{amsmath,amssymb,amsthm} % For including math equations, theorems, symbols, etc

\usepackage{varioref} % More descriptive referencing

\theoremstyle{definition} % Define theorem styles here based on the definition style (used for definitions and examples)

\theoremstyle{plain} % Define theorem styles here based on the plain style (used for theorems, lemmas, propositions)

\theoremstyle{remark} % Define theorem styles here based on the remark style (used for remarks and notes)

\hypersetup{
colorlinks=true, breaklinks=true, bookmarks=true,bookmarksnumbered,
urlcolor=webbrown, linkcolor=RoyalBlue, citecolor=webgreen, % Link colors
pdftitle={}, % PDF title
pdfauthor={\textcopyright}, % PDF Author
pdfsubject={}, % PDF Subject
pdfkeywords={}, % PDF Keywords
pdfcreator={pdfLaTeX}, % PDF Creator
pdfproducer={LaTeX with hyperref and ClassicThesis} % PDF producer
} % Include the structure.tex file which specified the document structure and layout

\title{Dual use issues in the field of\\Natural Language Generation} % The article title

\author{\spacedlowsmallcaps{Emiel van Miltenburg}\thanks{\textit{Department of Communication and Cognition, Tilburg University, Tilburg, the Netherlands.\\\textsc{siggen} board member from January 2021 to December 2024.\\Contact: \texttt{C.W.J.vanMiltenburg@tilburguniversity.edu}}}}

\date{} % An optional date to appear under the author(s)

\begin{document}

%----------------------------------------------------------------------------------------
%	HEADERS
%----------------------------------------------------------------------------------------

\renewcommand{\sectionmark}[1]{\markright{\spacedlowsmallcaps{#1}}} % The header for all pages (oneside) or for even pages (twoside)
\lehead{\mbox{\llap{\small\thepage\kern1em\color{halfgray} \vline}\color{halfgray}\hspace{0.5em}\rightmark\hfil}} % The header style

\pagestyle{scrheadings} % Enable the headers specified in this block

\maketitle % Print the title/author/date block

%----------------------------------------------------------------------------------------
%	ABSTRACT
%----------------------------------------------------------------------------------------

\section*{Abstract}
This report documents the results of a recent survey in the \textsc{siggen} community, focusing on Dual Use issues in Natural Language Generation (\textsc{nlg}). \textsc{siggen} is the Special Interest Group (\textsc{sig}) of the Association for Computational Linguistics (\textsc{acl}) for researchers working on \textsc{nlg}. The survey was prompted by the \textsc{acl} executive board, which asked all \textsc{sig}s to provide an overview of dual use issues within their respective subfields. The survey was sent out in October 2024 and the results were processed in January 2025. With 23 respondents, the survey is presumably not representative of all \textsc{siggen} members, but at least this document offers a helpful resource for future discussions. {\color{red} This report is open to feedback from the SIGGEN community. Let me know if you have any questions or comments!}

%----------------------------------------------------------------------------------------
%	TABLE OF CONTENTS & LISTS OF FIGURES AND TABLES
%----------------------------------------------------------------------------------------

\setcounter{tocdepth}{2} % Set the depth of the table of contents to show sections and subsections only

\tableofcontents % Print the table of contents

%\listoffigures % Print the list of figures

%\listoftables % Print the list of tables

%----------------------------------------------------------------------------------------
%	AUTHOR AFFILIATIONS
%----------------------------------------------------------------------------------------

%\let\thefootnote\relax\footnotetext{* \textit{Department of Communication and Cognition, Tilburg University, Tilburg, the Netherlands.\\\hphantom{* }SIGGEN board member from January 2021 to December 2024.}}

% \let\thefootnote\relax\footnotetext{\textsuperscript{1} \textit{Department of Chemistry, University of Examples, London, United Kingdom}}

%----------------------------------------------------------------------------------------

\newpage % Start the article content on the second page, remove this if you have a longer abstract that goes onto the second page

%----------------------------------------------------------------------------------------
%	INTRODUCTION
%----------------------------------------------------------------------------------------

\section{Introduction}
The term \emph{Dual use} has been introduced to the NLP community in an influential article by Dirk Hovy and Shannon L. Spruit \cite{hovy-spruit-2016-social}. They follow Jonas \cite{jonas1984imperative} in defining Dual Use problems as ''unintended consequences [of our research] that negatively affect people’s lives''\footnote{There are also other definitions in the NLP literature. For example, Kaffee and colleagues \cite{kaffee-etal-2023-thorny} define dual use as ``the intentional, harmful reuse of technology and scientific artefacts.'' This definition is narrower than the one offered by Hovy and Spruit. For this report, I will use the broader definition.} and go on to note that:

\begin{quote}
    ``We may not directly be held responsible for the unintended consequences of our research, but we can acknowledge the ways in which NLP can enable morally questionable/sensitive practices, raise awareness, and lead the discourse on it in an informed manner.''
\end{quote}

In the spirit of this quote, Vera Demberg (member of the \textsc{acl} executive board, responsible for \textsc{sig}s) reached out to all Special Interest Groups with the following four questions:

\begin{enumerate}
    \item What are the kinds of potential negative impacts of the technology area your SIG is focused on?

    \item How could your SIG take a role in raising awareness of both negative impacts and best practices for mitigating them among SIG members and also the broader community?

    \item Are there research topics in your SIG that might otherwise seem in scope but are either ethically or legally out of bounds?

    \item Relatedly, what known regulations apply to the application area your SIG focuses on, and how is the SIG going to help SIG members become aware of those regulations?
\end{enumerate}

As the \textsc{siggen} chair at that time, I decided it would be best to answer these questions through a survey among our membership. The \textsc{siggen} membership is defined as those who have signed up to the \textsc{siggen} mailing list (between 400 and 500 people).\footnote{The list is available at: \url{https://www.jiscmail.ac.uk/cgi-bin/webadmin?A0=SIGGEN}} The survey was approved by the other board members\footnote{Javier Gonzalez Corbelle, Raquel Herv\'as Ballesteros, David M. Howcroft, Chenghua Lin} and sent out in October 2024. This report summarizes the responses to our survey and discusses the implications of our findings.

\subsection*{Positionality statement}
This report aims to provide a factual summary of the responses to the survey, but this does not mean it is completely neutral. The structure of this document is influenced by my personal position, which I explain in the following. I believe that:
\begin{itemize}
    \item Dual use issues are real: our research may have unintended consequences.
    \item We may be able to avoid or mitigate some of these issues.
    \item Because some of these issues may be avoided or mitigated, we have a responsibility to learn about ways to achieve this.
\end{itemize}

This does not mean everyone should make ethics the focus of their research. Some people will, and some people won't. That is fine. But we \emph{should} try to stay up-to-date with the current best practices. I am aware that my position as a full-time academic puts me in a position of privilege, because I can actually spend time to do this. By writing this document, I hope to make this topic more accessible.
 
%----------------------------------------------------------------------------------------
%	METHODS
%----------------------------------------------------------------------------------------

\section{Approach}
\subsection{General design}
I implemented our survey through the Qualtrics survey platform. The four questions from the \textsc{acl} Executive board were used as the basis for the survey, and split into separate questions where necessary to obtain more detailed results. Once the questions were approved by the \textsc{siggen} board, the survey was distributed through the \textsc{siggen} list. All final questions are available in Appendix~\ref{sec:questions}.

\subsection{Target audience}
Our target audience consists of \textsc{siggen} members, who are both in academia or in industry.\footnote{Our main conference, \textsc{inlg}, has historically been visited by researchers and professionals alike. The attention for applications and practical issues is a great asset of our community.} Our community focuses on different aspects of language generation, ranging from linguistic studies of language production to the development of data-to-text systems. A common requirement in industry is that system outputs should be reliable, which means that rule-based approaches (or hybrid approaches combining rules with different kinds of machine learning algorithms) to \textsc{nlg} are still common. After all: Large Language Models still struggle to consistently produce factually correct output.

To make our survey as short as possible, there are no questions about demographic variables. This means that we cannot say anything about the difference between academia and industry in terms of their experience or stance regarding dual use issues. Given the exploratory nature of this survey, this is not a problem. First we need to identify which issues may be at play in our subfield, and in the future we can have further debates on how to deal with those issues.

\subsection{Data analysis}
I manually summarised the responses to each question of the survey. Similar answers are grouped together. Because open-ended questions tend to receive short answers, some interpretation was necessary to understand the motivation behind the answers. For example, if someone mentions a dual use issue, they may only briefly describe the use, but not what the negative effects are of this use of technology. In such cases, I have tried to elaborate on what the respondents may have meant to clarify the issue.

A final disclaimer: not all responses may be seen as \emph{proper} dual use issues; I have tried to find a balance between (1)~critically assessing the relevance of each response and (2)~being as inclusive as possible to be able to address everyone's concerns.

\subsection{Ethical considerations}
In earlier work, I conducted another survey among the \textsc{siggen} members \cite{van-miltenburg-etal-2023-barriers}, for which I obtained ethical approval. The current survey uses the same approach: participants first read an information letter (Appendix~\ref{sec:informationletter}), after which they agree to the conditions of the survey on a separate informed consent page (Appendix~\ref{sec:informedconsent}). The survey is conducted fully anonymously, and the option to collect IP addresses of our respondents has been disabled in Qualtrics (this feature is normally turned on by default). Whereas the previous survey asked participants about their level of experience (in different ranges of years) and their general affiliation (industry or academia), this time we opted to leave out any personal information and only ask participants to respond to our main questions. Participation is completely voluntary, and participants are offered no compensation for their efforts (aside from receiving this report through the \textsc{siggen} list).

%----------------------------------------------------------------------------------------
%	RESULTS AND DISCUSSION
%----------------------------------------------------------------------------------------

\section{Results}
In total, 23 \textsc{siggen} members filled in our survey. The six subsections below provide a summary of their responses.

\subsection{Adverse effects of \textsc{nlg} technology}
What dual use issues are \textsc{siggen} members aware of? I have listed the main issues below, with sub-lists indicating further examples that roughly fall in the same category. No counts are provided, since this is a qualitative study aiming to describe what issues people are aware of. 

\begin{itemize}
    \item \textsc{nlg} technology can be used to create texts that elicit behaviour change. Persuasive technology can be used for good (e.g.\ to help people quit smoking or to stick to a healthy diet) but also to induce undesirable behaviour.
    
    \item \textsc{nlg} technology can be used to reduce or replace human interaction. This can reduce friction and make certain tasks more efficient, but makes it harder for people to interact with other human beings. For example, patients may not be able to talk to a real doctor anymore. Other examples:
    \begin{itemize}
        \item One respondent mentioned AI companions replacing human relationships. They did not explain why this is a dual use issue, but perhaps the problem is that the use of AI companions may lead to further social isolation.

        \item Two participants mentioned bots on social media that pollute online communities and make it harder to find sincere human interactions.
    \end{itemize}

    \item \textsc{nlg} technology may be used to create misinformation or disinformation (also known as `fake news') at a large scale. This undermines democracy, where citizens are meant to stay informed so they can make well-founded decisions.

    \item \textsc{nlg} technology may be used to non-consensually impersonate other people. These impersonations are also known as `deepfakes' although this term is most often used for images or videos.

    \begin{itemize}
        \item Three respondents mentioned that the use of \textsc{nlg} technology may lead to an increase of scamming or phishing practices. These practices may involve impersonation, but scammers could also `only' use \textsc{nlg} technology to personalise their messages without impersonating others. One participant also mentioned that recent improvements in text-to-speech may enable spam that is delivered via the telephone.
    \end{itemize}

    \item \textsc{nlg} technology may be used to replace humans in the creative industry. This means that people who enjoy their creative work may lose their livelihood in favor of an \textsc{nlg} system.
    \begin{itemize}
        \item One participant mentioned that this may happen indirectly as well: books or websites with automatically generated texts (that are basically paraphrased from other sources) may take away revenue from books or websites with human-authored texts. By using \textsc{nlg} technology, copyright restrictions may be circumvented, which makes it harder to address this issue.
    \end{itemize}

    \item The use \textsc{nlg} technology may bring about privacy or security risks, if either the training data or the real-world interactions with \textsc{nlg} systems leak information to other users.

    \item \textsc{nlg} technology may be used to undermine academic integrity. Authors may choose to automatically generate texts instead of writing their own. In the case of students using \textsc{nlg} to carry out their writing assignments, this leads to unreliable assessments of students' true capabilities. In the case of researchers, this may lead to an overburdening of the publication system (in the case of automatically generated manuscripts) or unreliable reviews (in the case of automatically generated reviews).
    \begin{itemize}
        \item One participant specifically mentioned the lack of proper attribution as an issue here. Since we do not know what sources language models' outputs are based on (and solutions to address this issue are still under development), original authors may not be credited for their work.
    \end{itemize}

    \item \textsc{nlg} technology may be used to generate an overabundance of online content, making it harder to find useful and original information. Moreover, these generated texts could be optimised to be highly ranked by different search engines. This has negative effects on our information ecosystem.

    \item The use of \textsc{nlg} technology may lead to an over-reliance on automatically written texts. Because current models may still produce erroneous or biased output, this may lead to poor decisions or the proliferation of biases in society.
    \begin{itemize}
        \item One participant notes in this context that the human-likeness of conversational interfaces may contribute to the problem of users overly relying on automatically generated responses.

        \item Another participant notes that the use of AI-generated summaries may result in people losing their ability to critically evaluate longer or more difficult texts.

        \item A different respondent points to the literature on LLM safety for more discussion of potential negative effects. In medical settings, the worst-case scenario is that patients die because of poorly informed decisions.
    \end{itemize}

    \item The use of \textsc{nlg} technology may lead to a loss of language diversity, as texts become more homogeneous due to them being produced by the same (small) set of language models.

    \item The use of \textsc{nlg} technology may lead to environmental harm, due to the energy consumption of current models and the water use of data centers.

    \item The use of \textsc{nlg} technology may further reinforce the dominant position of a select number of majority languages that are supported by current technological solutions. This may have a detrimental effect on minority languages that are not supported.

    \item A final issue concerns the way that \textsc{nlg} systems are designed. If we develop systems that are not transparent in their decision-making, then it may not be possible for stakeholders to properly assess the decisions made by those systems. This response seems to appeal to the \emph{right to explanation} \cite{vredenburgh2022right}, i.e.\ the right to know why a particular decision was made if that decision affects your everyday life.
\end{itemize}

\subsubsection*{Other ethical issues}

One response mentioned the use of copyrighted texts without permission to train Large Language Models as a dual use issue. Though the effects of using copyrighted texts may well fall into this category, the use of these texts alone is not strictly speaking a dual use issue, as it is not an \emph{unintended consequence} of using \textsc{nlg} technology but instead it is a potential cause of dual use issues, and a practice of which the ethical and legal status is \emph{directly} being contested.

Another response mentioned the impacts of current crowdsourcing practices on the mental health of the crowdworkers who take part in the training process of large language models by providing feedback on the output of these models. It is by now well known that many of these workers are underpaid and that they have little to no mental health support. This again does not seem to be a dual use issue, since it is clearly in our control to treat crowd workers better. Calling this a dual use issue would reduce our responsibility to avoid harming any stakeholders in the research process.

\subsubsection*{On the rise of dual use issues}
Another respondent attributed many dual use issues to the rise of Large Language Models, because they suffer from hallucination issues where these do not occur with rule-based systems. A further difference between earlier \textsc{nlg} systems and LLMs is that the former are also much less resource-intensive, leading to fewer environmental concerns.

\subsubsection*{The point of this survey}
Two respondents did not seem to see the point of this survey:

\begin{quote}
    ''Like any other technology, it could be use[d] in a malicious way, but I don't see the specificity to \textsc{nlg} technology.'' \hfill (Participant 9)\\

    ``I'm sorry but I think this is a ridiculous question. Which potential negative effects are you aware of matches, a hammer, a kitchen sink, or any implement depends on how creative you want to get.''\\
    \null\hfill (Participant 15)
\end{quote}

While it is true that anything can be used in a nefarious way, it is also important to be mindful of our privileged position in society as academics who are paid to acquire knowledge and to share our findings. To quote Stan Lee's famous adage: \emph{with great power comes great responsibility.} Given our position, we should ask ourselves (1) what kind of society do we want to live in? (Or in other words: what values do we hold dear?) and (2) how do our actions influence society? Given the issues that may arise from the technology that we develop, it is worth thinking about ways in which these issues could be avoided.

\subsection{Mitigating dual use issues}
How can we mitigate dual use issues? I have categorised the responses into four different categories, which will be discussed in turn below:
\begin{enumerate}
    \item Raising awareness
    \item Research
    \item Technical solutions
    \item Regulation and policy (also explicitly addressed later in the survey)
\end{enumerate}

\subsubsection*{Raising awareness}
The first category of responses is all about raising awareness: making sure that users are aware of the risks and limitations of language technology, and making sure that researchers are aware of potential dual use issues. For researchers, this awareness could be stimulated by:

\begin{itemize}
    \item Organising workshops
    \item Publishing reports (such as this one)
    \item Actively requiring authors to think about dual use issues in their work.
\end{itemize}

\noindent For users, this awareness could be stimulated through:
\begin{itemize}
    \item Educating people on the limitations and risks of language technology, and how to critically assess automatically generated content.
    \item Giving people practical advice for when language technology can and cannot be used.
    \item Other forms of public outreach by \textsc{nlg} researchers.
\end{itemize}

An important note is that users should not just be taught to be aware of potential issues with language technology, but also how to navigate those issues.

\subsubsection*{Research}
The classical academic response to any question: \emph{more research is needed} (closely followed by the other classical response: \emph{it depends}). Jokes aside, different respondents underlined the need to deepen our understanding in different areas. Suggested topics are:

\begin{itemize}
    \item To better understand the causes and effects of model bias and hallucinations.
    \item To better understand how people use LLMs exactly, and how these models affect their users.
    \item To explore alternative approaches to \textsc{nlg} that do not require scraping of online data or modeling online text.
    \item To explore better output evaluation and validation techniques.
    \item To further discuss and explore mitigation strategies.
\end{itemize}

Some other research ideas are presented under the header of technological solutions below.

\subsubsection*{Technical solutions}
Different technical solutions were mentioned to address different dual use issues.

\begin{itemize}
    \item We might be able to build systems that are able to detect automatically generated text, or texts that are otherwise malicious/unsafe.

    \item We might be able to develop better methods for data curation.
    \begin{itemize}
        \item One respondent suggested that this could reduce the toxicity of the training data.
        \item Another responded suggested that this could help to improve the diversity of the training data, so that minorities/underrepresented groups are also included.
    \end{itemize}

    \item We might be able to insert more knowledge into current \textsc{nlg} models.

    \item We might be able to develop methods to make the reasoning process behind the outputs more transparent.
    \begin{itemize}
        \item One respondent mentioned chain-of-thought reasoning as a potential solution.
    \end{itemize}

    \item We might be able to use text classifiers to help LLMs detect both malicious/unsafe inputs and outputs.

    \item We might be able to develop more energy- and resource efficient models.
    
    \item We might be able to put guard rails on Large Language Models so that (1) they do not produce undesirable outputs, and that (2) they refuse to carry out particular tasks.

    \item We might be able to develop methods to enforce safety guarantees, such as fidelity to the input (or general reference) data.

    \item We might be able to develop models that can accurately identify the sources that automatically generated texts are based on.
    \begin{itemize}
        \item Different respondents identified Retrieval-Augmented Generation (\textsc{rag}) as a potential solution.
    \end{itemize}

    \item We might be able to watermark the outputs of \textsc{nlg} systems so that they are easily recognisable as such.
    \begin{itemize}
        \item One respondent suggested that watermarking schemes should be made public, so that others can easily recognise through which model texts were automatically generated. Watermarking can only work if all or at least the majority of available models use this technique.
    \end{itemize}
    
\end{itemize}

At this point I have to note that many of these solutions are already being explored, and that technological solutions to dual use issues may themselves may also have limitations and unintended consequences. For example: automatic detection of automatically generated essays could in theory be a useful tool to counter students who hand in automatically generated essays. But what about the students who are wrongfully accused of submitting automatically generated essays? One has to wonder whether technological solutions are the right approach to handle social issues. (Why are students submitting automatically generated essays in the first place?) As the reader will hopefully understand, dual issues for technological solutions to dual use issues (and solutions to those meta-dual use issues) were beyond the scope of this survey.

\subsubsection*{Regulation and policy}
Our respondents highlighted different forms of regulation that could be introduced at five different levels: university, government, research community, professional community, website registries. All suggestions are provided below. Some of these have already been developed or implemented, other suggestions are yet to be taken up.

\begin{itemize}
    \item Open source as much as possible, so that others may critically assess your work.
    \item Mandate clearly visible labels for automatically generated content.
    \item Mandate the use of model cards to explicitly describe each model's strengths and weaknesses.
    \item Mandate the use of data statements to explicitly describe the data that each model was trained on.
    \item Encourage the involvement of different stakeholders in each stage of model development.
    \item Develop (and potentially enforce) ethical guidelines for developers.
    \item Require authors to add limitations sections to their work, where they explain the risks of their work.
    \item Require authors to have their research proposals assessed by ethical review boards.
    \item Governments can develop regulations to counter particular dual use issues.
    \item Website registries could develop enforceable guidelines for websites, so that content farms can no longer be set up without any consequences.
\end{itemize}

\subsubsection*{Further observations}
One respondent noted as a mitigation strategy that:

\begin{quote}
    ``People who don't want to use AI are not forced to do this.''\\\null \hfill(Participant 0)
\end{quote}

The problem with this statement is that people who do not use AI may still be affected by those who do. Virtually all dual issues (except perhaps increased social isolation) are the result of \emph{others} using \textsc{nlg} technology.

\subsubsection*{Future work}
I have not attempted to create a mapping between all dual use issues and all mitigation strategies, since creating such a mapping is a huge undertaking. It would probably be a good idea to first explore the different dual use issues in more depth to understand if and how they are all connected to each other, and to then try to create an overview of the mitigation strategies for each `dual use cluster.'

\subsection{Raising awareness of dual use issues and mitigation strategies}\label{sec:aware-of-issues}
How can \textsc{siggen} help to raise awareness of both dual use issues and mitigation strategies? Different respondents suggested different audiences: researchers, practitioners, the general audience, and even policy makers such as politicians and governments. So next to the question of \emph{how} we should raise awareness, there is also the question of \emph{who} the target of those efforts should be. This question is left for future discussions.

\begin{itemize}
    \item Disseminating this report and potential follow-up reports, perhaps including guidelines for \textsc{nlg} researchers. These could also be published on the \textsc{siggen} website.
    \item Running webinars or other online events about dual use issues and mitigation strategies.
    \item Making space to discuss dual use issues at the \textsc{inlg} conference.
    \begin{itemize}
        \item Create a separate track at \textsc{inlg} to discuss dual use and other ethics issues. Submissions in this track could also be eligible for awards.
        \item Organise panels or birds-of-a-feather sessions on the topic of ethics and dual use.
        \item Organise shared tasks to mitigate particular dual use issues.
    \end{itemize}
    \item Improving review quality, or at least putting more emphasis on different aspects of \textsc{nlg} papers that we find to be particularly important.\footnote{A personal observation is that the quality of the reviews at \textsc{inlg} is typically better than that of larger conferences such as \textsc{acl}. Reviewers are typically more knowledgeable, because they have all been selected from the \textsc{nlg} community, and they are typically more constructive. (Perhaps because perceived prestige is less important at smaller conferences. Everyone is just excited about the topics that other people in their research area are working on).}
    \item Organise courses or workshops on \textsc{nlg} safety. One respondent noted that we may wish to target younger audiences so that the next generation of \textsc{nlg} researchers and practitioners will be aware of dual use issues and mitigation strategies.
    \item Get in touch with industries that are set to be replaced with generative AI tools, and give a platform to stakeholders who experience the impact of \textsc{nlg} technology. Examples could be journalists and people in the creative industry.

\end{itemize}

Multiple participants noted that \textsc{siggen} is a relatively small community, and that it may be wise to partner with other (sub)communities, or to let \textsc{acl} take the lead on dual use issues.

\subsubsection*{Who should be responsible for this?}
The \textsc{siggen} board is a relatively small group of people who help support the community by staying in contact with \textsc{acl}, maintaining the website, keeping an eye on the finances and the mailing list, and looking for people willing to organize \textsc{inlg}. Should the same group of people be responsible for developing resources on dual use issues? Perhaps it is better to set up a separate working group to exclusively focus on the ethics of \textsc{nlg} research. Or perhaps we should just encourage and make space for initiatives in this area, without appointing a select group of people to work on this topic.

\subsubsection*{Who can \textsc{siggen} represent?}
One respondent noted that \textsc{siggen} could not just raise awareness among \textsc{nlg} researchers, but that we as a community could also reach out to different government officials to voice our concerns. An issue with this suggestion is the idea of speaking in the name of all \textsc{nlg} researchers. Not all members of our community share the same concerns. 

\subsection{Unethical or illegal research topics}
Are there topics that might otherwise seem in scope for \textsc{nlg} research, but are either ethically or legally out of bounds? Only seven people answered \emph{yes} to this question. Topics that were mentioned are:

\begin{itemize}
    \item Style transfer of copyright protected authors.
    \item Stimulation of parasocial relationships between humans and AI companions.
    \item Other forms of impersonation.
    \item Creation of persuasive text for manipulation.
    \item Generating malicious content.
    \item Military applications.
\end{itemize}

I should note that it is possible to find good reasons to study each of these topics, but researchers should be aware that these are at the very least sensitive topics that need to be handled with care.

\subsection{Known regulations}
What known regulations apply \textsc{nlg} research or \textsc{nlg} technology? Our respondents identified six different sets of regulation:

\begin{enumerate}
    \item The European AI act.
    \item The General Data Protection Regulation (GDPR) in Europe.
    \item Ethical requirements from funding agencies.
    \item Academic regulations about the protection of human subjects.
    \item Regulations for medical devices.
    \item Copyright, liability and defamation laws.
\end{enumerate}

\subsection{Raising awareness of regulations}
How can \textsc{siggen} help raise awareness of current regulations? This question is similar to the question about awareness of dual use issues and mitigation strategies (\S\ref{sec:aware-of-issues}), but since the topic of regulations is much more narrow, there are also fewer possibilities to raise awareness.

\begin{itemize}
    \item Creating an overview of the different regulations on the \textsc{siggen} website.
    \item Organise invited talks/panels/tutorials on the topic of regulation. These could take place online, or at physical conferences such as \textsc{inlg}. Once such an event has taken place, a recording of the event could also be put online.
    \item Explicitly refer to existing regulations in calls for papers and in the reviewing process.
    \item Keep track of the attention paid to dual use in the community, and write reports on this topic.
\end{itemize}

Two potential issues were raised by the respondents. The first issue is that existing regulations may be complex and thus hard to understand. So awareness is not enough; the regulations should also be accessible. The second issue is that community members who come from industry may not have access to ethical review boards. If our conferences become more strict in enforcing particular regulations, we have to make sure that it is feasible for everyone to comply.

%----------------------------------------------------------------------------------------
%	Concluding remarks
%----------------------------------------------------------------------------------------

\section{Concluding remarks}
This report aims to provide an overview of the dual use issues that different \textsc{siggen} researchers are aware of, and to enable further discussions on this topic in our community. Of course much more could be, and indeed has been said about dual use issues \cite[for example]{kaffee-etal-2023-thorny,bender-etal-2020-integrating,brenneis2024assessing,grinbaum2024dual,leins-etal-2020-give}. For this report, I have chosen to focus on the responses provided in our survey, and to not provide a full overview of the literature on dual use issues. This is mostly for reasons of time: I wanted this report to be available as soon as possible, and it would have been irresponsible to give a quick-and-dirty overview without really doing justice to the field.

\subsection{Disclaimer: this report contains personal opinions}
I would like to reiterate that this report is a summary of the responses provided by members of the \textsc{siggen} community, summarised by another community member. I thought it might be useful to add my own thoughts on dual use as a way to make this document more readable (not just a list of responses) and to stimulate discussion (by having something to agree or disagree with). I have not consulted with the \textsc{siggen} board to determine the official \textsc{siggen} stance on this topic, so this document may not be reflective of the board's opinion. In other words: the commentary in this document reflects my personal stance.

\subsection{Disclaimer: \emph{not all \textsc{nlg} research}}
An easy response to this report would be: not all \textsc{nlg} research suffers from these issues! I fully agree. Different kinds of \textsc{nlg} systems and different research topics have different issues that are associated with them. But as a community, we do have a collective responsibility to hold each other accountable for our work and to reflect on our role in the ecosystem. Of course, there is much to be said about the ways in which we can hold each other accountable in a respectful and inclusive manner.

\subsection{Next steps}
What are the next steps for us as a research community? Having established a list of potential issues and mitigation strategies through this survey, one might organise focus groups to have a nuanced discussion about the ways in which both individual researchers and the research community as a whole should handle dual use issues. Given the size of the list, it is not feasible to cover everything in one session and `get it over with.' Dual use issues require a \emph{sustained effort} from our community to minimise the harms that may arise from our research. 

\section*{Acknowledgments}
I would like to thank Vera Demberg and the \textsc{acl} Executive board for their questions. Thanks to the \textsc{siggen} board for their feedback on the questions for the survey, and to the \textsc{siggen} community for their responses.

%----------------------------------------------------------------------------------------
%	BIBLIOGRAPHY
%----------------------------------------------------------------------------------------

\renewcommand{\refname}{\spacedlowsmallcaps{References}} % For modifying the bibliography heading

\bibliographystyle{unsrt}

\bibliography{sample.bib} % The file containing the bibliography

%----------------------------------------------------------------------------------------

\part*{\Large Appendices}

\appendix

\section{Invitation email}
Dear SIGGEN members,\\

\noindent I am sending this message on behalf of the SIGGEN board. We have been asked by the ACL to look into potential dual use issues within our area of interest: Natural Language Generation (NLG). The term Dual use refers to potential negative uses of (the products of) our research. To get a better sense of the potential impact of both NLG research and NLG technology, we are asking all members to fill in this short (5-10 minutes) survey: URL.\\

\noindent Best wishes,

\noindent Emiel van Miltenburg 

\section{Information letter}\label{sec:informationletter}
Dear SIGGEN members,\\

We have been asked by the ACL to look into potential dual use issues within our area of interest: Natural Language Generation (NLG). The term Dual use refers to potential negative uses of (the products of) our research. To get a better sense of the potential impact of both NLG research and NLG technology, we are asking all members to fill in this survey.

Through this survey, we hope to be able to answer the following four questions (as formulated by Vera Demberg, ACL exec member responsible for SIGs):
\begin{enumerate}
    \item What are the kinds of potential negative impacts of the technology area your SIG is focused on? [SIG stands for Special Interest Group, in our case: SIGGEN]
    \item How could your SIG take a role in raising awareness of both negative impacts and best practices for mitigating them among SIG members and also the broader community?
    \item Are there research topics in your SIG that might otherwise seem in scope but are either ethically or legally out of bounds?
    \item Relatedly, what known regulations apply to the application area your SIG focuses on, and how is the SIG going to help SIG members become aware of those regulations?
\end{enumerate}

We believe that part of the answer to the second and fourth question is actually to distribute this survey and report the results in the first place. But we are very interested to hear your views as well!    Duration of this survey  We aimed to keep this survey as short as possible, and roughly expect that filling in the survey should take about 5-10 minutes.

\paragraph{Potential benefits and negative consequences}

We foresee no negative consequences of filling in this survey. Your responses to this survey are fully anonymous, and thus cannot be traced back to you. Our community as a whole will benefit from your answers, since we will have a better picture of the challenges that different researchers are facing in this area. We aim to summarise our findings in a report shared with the SIGGEN community.

\paragraph{Voluntary participation}

Participation is voluntary, without any financial compensation. You are under no obligation to complete the survey, and you may decide to quit the survey at any point, for whatever reason, without any negative consequences.    

\paragraph{What will happen with the data?}

As noted above, we (the SIGGEN board) will write a report summarising the results of the survey. We will publish the results through ArXiv for future reference, and share a link to the survey through the SIGGEN list. We aim to be maximally transparent, and as such we will also publish the survey responses so that others may carry out additional analyses of the responses. If you have any concerns about your responses being made public, you are also free to message the board directly instead of responding through this survey.

\paragraph{Who is responsible for this survey?}

This survey has been created by Emiel van Miltenburg, and has been vetted by the rest of the SIGGEN board (Raquel Hervás, David Howcroft, Chenghua Lin, Javier González Corbelle). Although this survey has not been approved by any ethics committee, it does follow common standards and best practices.

If you have any questions, or if you find any issues with this survey, please feel free to contact the author of this survey at: C.W.J.vanMiltenburg@tilburguniversity.edu. You can also contact the SIGGEN board at: siggenboard@googlegroups.com.

\section{Informed consent form}\label{sec:informedconsent}
By clicking "yes" below:
\begin{itemize}
    \item You confirm that you have read the information letter on the previous page. 	You confirm that you have been able to ask any questions (via email) before taking part in this survey.
    \item You consent to your responses being used to write a report on dual use issues related to Natural Language Generation.
    \item You acknowledge that the report along with all responses to this survey will be made public.
    \item You agree to take part in this survey.
    \item You acknowledge that your participation is voluntary, and that you may quit at any time, without any negative consequences. 
\end{itemize}

Answer options:
\begin{itemize}
    \item Yes, I consent
    \item No, I do not consent \hfill $\rightarrow$ Skip to end of survey
\end{itemize}

\section{Questions used in the survey}\label{sec:questions}
These questions were based on the four initial questions by Vera Demberg, but some questions were edited into multiple questions to obtain better answers.
\begin{enumerate}
    \item What potential negative effects of NLG technology are you aware of?  Please use newlines (ENTER) to separate your answers. (Open question)
    \item What best practices to mitigate the potential negative effects of NLG technology are you aware of? Please use newlines (ENTER) to separate your answers. (Open question) 
    \item How can SIGGEN play a role in raising awareness of these two? That is:
    (1) Potential negative effects of NLG technology (2) Ways to mitigate these negative effects? Please use newlines (ENTER) to separate your answers. (Open question)
    \item Are there topics that might otherwise seem in scope for NLG research, but are either ethically or legally out of bounds? (Answer options: No, Maybe/unsure, Yes)
o	\begin{description}
    \item[if Yes:] Can you provide examples of topics that might otherwise seem in scope for NLG research, but are either ethically or legally out of bounds?  Please use newlines (ENTER) to separate your answers. (Open question)
\end{description}
    \item What known regulations apply NLG research or NLG technology?  Please use newlines (ENTER) to separate your answers. (Open question)
    \item How can SIGGEN play a role in raising awareness of those regulations?  Please use newlines (ENTER) to separate your answers. (Open question)
\end{enumerate}

\end{document}